\documentclass[10pt,conference]{IEEEtran}
\IEEEoverridecommandlockouts

\usepackage{cite}
\usepackage{amsmath,amssymb,amsfonts}
\usepackage{algorithmic}
\usepackage{graphicx}
\usepackage{textcomp}
\usepackage{xcolor}
\def\BibTeX{{\rm B\kern-.05em{\sc i\kern-.025em b}\kern-.08em
    T\kern-.1667em\lower.7ex\hbox{E}\kern-.125emX}}
\begin{document}

\title{Rethinking Data: Towards Better Performing Domain-Specific Small Language Models}

\author{Boris Nazarov \qquad Darya Frolova \qquad
Yackov Lubarsky \qquad Alexei Gaissinski \qquad Pavel Kisilev
\\Huawei, Israel}

\maketitle

\begin{abstract}
Fine-tuning of Large Language Models (LLMs) for downstream tasks, performed on domain-specific data has shown significant promise. However, commercial use of such LLMs is limited by the high computational cost required for their deployment at scale. On the other hand, small Language Models (LMs) are much more cost effective but have subpar performance in a similar setup. This paper presents our approach to fine-tuning a small LM, that reaches 
high accuracy in multiple-choice question answering task. We achieve this by improving data quality at each stage of the LM training pipeline.  In particular, we start with data structuring resulting in extraction of compact, semantically meaningful text chunks used by a retriever. This allows more efficient knowledge digestion by the LM. Further, we improve the retrieved context by training a lightweight Chunk Re-Ranker (CRR) that generates more accurate relative relevance chunk scores. Finally, we improve the model generalization ability by merging the models fine-tuned with different parameters on different data subsets. We present detailed procedure descriptions, and corresponding  experimental findings that show the improvements of each one of the proposed techniques. 

\end{abstract}


\section{Introduction}
 Multiple choice question answering (MC-QA) is a standard important task used for evaluation of Language Models (LMs). This paper presents our methodology of optimizing a small LM for MC-QA for a specific domain, with a case study in the field of telecommunications. 
 Our work underscores the importance of targeted data analysis and preparation at each stage - from initial data pre-processing to accurate selection of the context for LLM prompting. This rigorous data analysis combined with advanced fine-tuning and retrieval techniques boosts model performance on domain specific data, and brings substantial improvements in accuracy and efficiency.
 
This work was done under the 2024 \textit{Specializing Large Language Models for Telecom Networks by ITU AI/ML in 5G Challenge}, organized by \cite{b1}. The competition aimed to exploit the potential of LLMs in the telecommunications industry. The goal was to design resource-efficient domain-specific LLMs by developing novel approaches or combining existing methods, such as fine-tuning, Retrieval Augmented Generation (RAG) and prompt engineering. Participants were limited to two models of different sizes: Falcon-7.5B \cite{b2} and Phi-2 \cite{b3}. The competition task was to achieve the highest accuracy by improving the baseline model, while answering a set of multiple-choice questions from the TeleQnA dataset \cite{b14}. To account for the size differences between Falcon and Phi-2, Falcon-based solutions were prohibited from fine-tuning the model. The challenge data was limited to a set of TeleQnA multiple-choice questions related to different classes of telecom knowledge domains, and to the corresponding documents.

We selected the Phi-2 model as our base model, as we aimed to demonstrate that strong results can be achieved even with a small model and a limited amount of data. We aimed to implement an efficient pipeline, integrating data pre-processing, fine-tuning and advanced retrieval techniques, in order to achieve the highest possible accuracy. 

\textbf{Summary of contributions:}
\begin{itemize}
\item \textit{Data representation.} We structure unprocessed data in order to allow better knowledge digestion by the small model (2.5B in our experiments). Specifically, we create an automatic standardized data restructuring pipeline. It allows to extract semantically meaningful text chunks of minimal size that retain maximal information for the question answering. This allows the retrieval system to access the smallest possible number of chunks, thereby minimizing the context length, while ensuring sufficiently comprehensive information to extract accurate answers.

\item \textit{Chunk Re-Ranking} In order to improve the retrieved context relevance to the question, we train the Chunk Re-Ranker (CRR), based on the same Phi-2 model. It is trained to assign a
\textit{relative relevance score} to each retrieved chunk. We use it together with the score provided by the Retriever, to create the context out of the newly re-ranked and filtered chunks. This allows us to retain the most relevant context information pertaining to the provided question.

\item \textit{Fine-tuning and models merging.} Enhanced model robustness and generalization achieved through the merging of models that are fine-tuned with different input parameters on different data subsets.
\end{itemize}

\section{Previous Work}

\subsection{Multiple choice question answering (MC-QA)}\label{Prev_MCQA}

\begin{figure}
    \centering
    \includegraphics[width=0.9\linewidth]{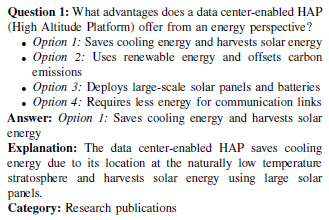}
    \caption{Data example from TeleQnA dataset \cite{b14}. Each data unit is represented in standardized format, comprising of five distinct fields: [Question, Options, Answer, Explanation, Category]}
    \label{fig:data_example}
\end{figure}

Despite the significant success of LLMs in the NLP tasks, their comprehensive evaluation remains an inevitable challenge. Among other LLM benchmarks the utilization of Multiple Choice Question Answering (MC-QA) has gained extreme popularity. 
MC-QA task is an automatic, non-subjective evaluation method with accuracy as resulting metric, widely applied for numerous LLMs to test both their commonsense knowledge and domain-specific knowledge \cite{b10}, \cite{b11}, \cite{b12}. Although the rationality of MC-QA as evaluation method for LLMs is still under active discussion \cite{b6}, it became and remains to be a fundamental format for various tasks in NLP (e.g. \cite{b7}, \cite{b8}, \cite{b9} etc). In MC-QA each entry contains a question associated with several answer options, and the model is tasked to identify the correct response. The example on Fig.\ref{fig:data_example} contains 2 additional fields - 'Explanation' and 'Category', that are not necessarily presented in the MC-QA data, but may be helpful for model fine-tuning, if provided.

Despite the intuitiveness of the MC-QA itself for LLM evaluation, creating data for it is a challenging task, and existing MC-QA datasets are available for limited domains. The commonly used leaderboards designed to track progress in MC-QA can be found in \cite{b13} with 8 datasets of different domain specification. Also the researchers keep creating more specific and challenging benchmark datasets designed to evaluate LLMs in various technical fields (\cite{b15}, \cite{b14}).

\subsection{Data pre-processing}\label{Prev_DataPrep}

The data quality can significantly impact the capacity and performance of the LLM, that is why a lot of effort is put to pre-process the collected text data to remove noise, redundancy, irrelevance, and potentially harmful content. The general aspects of high-quality training data were addressed in \cite{b5}. The authors analyzed how Extractive QnA (the task of answering questions given a context \cite{b4}) actually depends on the high quality training data, and even investigated the possibility of unsupervised Extractive QnA. 

The important pre-processing strategy to enhance the collected data's quality is data deduplication. In \cite{b16} the authors showed that LLMs trained on deduplicated data memorize less of their data and require fewer training steps to achieve the same or better accuracy. Many researches followed the advice to deduplicate their training data to achieve better results \cite{b17}.

\subsection{Prompting}\label{Prev_Prompt} 

It is hard to overestimate the impact of the prompting on LLM performance in MC-QA \cite{b18}. For MC-QA two major types of prompts should be mentioned: cloze prompting and multiple choice prompting (MCP). In cloze prompting, a question is passed through the model, the candidate answers are each independently scored by LLM, and the answer option assigned the highest probability is chosen to be a correct answer. On the other hand, in MCP a question and its candidate answers are all passed through the model in a single prompt. The prompt is structured such that the LLM must only predict a single token (such as “A”, “B” or "1", "2" etc.). The token assigned the highest probability is associated with the answer choice, and is chosen to be the model’s answer. There are several problems with cloze prompting that do not apply to multiple choice prompting, like expense, dependency on the normalization procedure, the fact that only implicit candidate answers comparison through their final probabilistic scores etc. The authors in \cite{b18} demonstrated that when LLM is prompted MCP instead of CP, performance often dramatically improved – approaching or even surpassing SOTA performance on a variety of datasets (16 out of 20 datasets).

\subsection{RAG-based approaches}\label{Prev_RAG}

One of the most common use-cases for LLMs for MC-QA task is to provide an answer over a set of data. These Retrieval-Augmented Generation (RAG) models \cite{b29} are a potential solution to the problems of generating incorrect or not up-to-date information and hallucinations. In \cite{b30} the authors introduced PaperQA - an agent that performs information retrieval across full-text scientific articles and uses RAG to provide answers.
The authors in \cite{b28} proposed compressing the retrieved documents into textual summaries prior to in-context integration in order to reduce the computational costs and to improve the identification of relevant information in long documents.
Another way to improve the quality of data retrieval is to use a reranking model. This model provides a relevance score given a query and document pair. The score is used to reorder and filter the documents by relevance to our query. Wide range of reranking methods: from \cite{b35} to G-RAG, a reranker based on graph neural networks \cite{b31}

\subsection{Models fusion}\label{Prev_Merging}

Model merging (see overview in \cite{b19}) is a technique to create a single powerful model from multiple LLMs. A created model should exhibit a broader range of capabilities and show improved overall performance.

Model Soups paper \cite{b20} proposed combining multiple models by averaging their weights (a "linear" merging). 

\cite{b21} also merged weights from models with identical structures to achieve improved overall performance.

There have been a few attempts to enhance out-of-distribution model generalization and generalization over domains and tasks via weighted averaging of models obtained with different configurations: \cite{b22}, \cite{b23}, \cite{b24}, \cite{b25}. In \cite{b26}, \cite{b27} authors showed how linear mathematical operations can be applied also to adapter parameters to achieve higher generalization performance.

\section{Our Approach}
Our objective was to develop an effective method for integrating and deploying a \textit{small} language model into the emerging field of telecommunications. The smaller size allows the model to be more efficient in many aspects than its larger competitor. However, since in LMs a model capacity has been shown to correlate with the its size, small LMs achieve lower overall capabilities. The challenge of fine-tuning a small LM is thus to determine optimal model and input data characteristics - a trade-off between model size, performance, customizability on the one side, and amount and quality of the training data on the other side.

\begin{figure}
    \centering
    \includegraphics[width=1\linewidth]{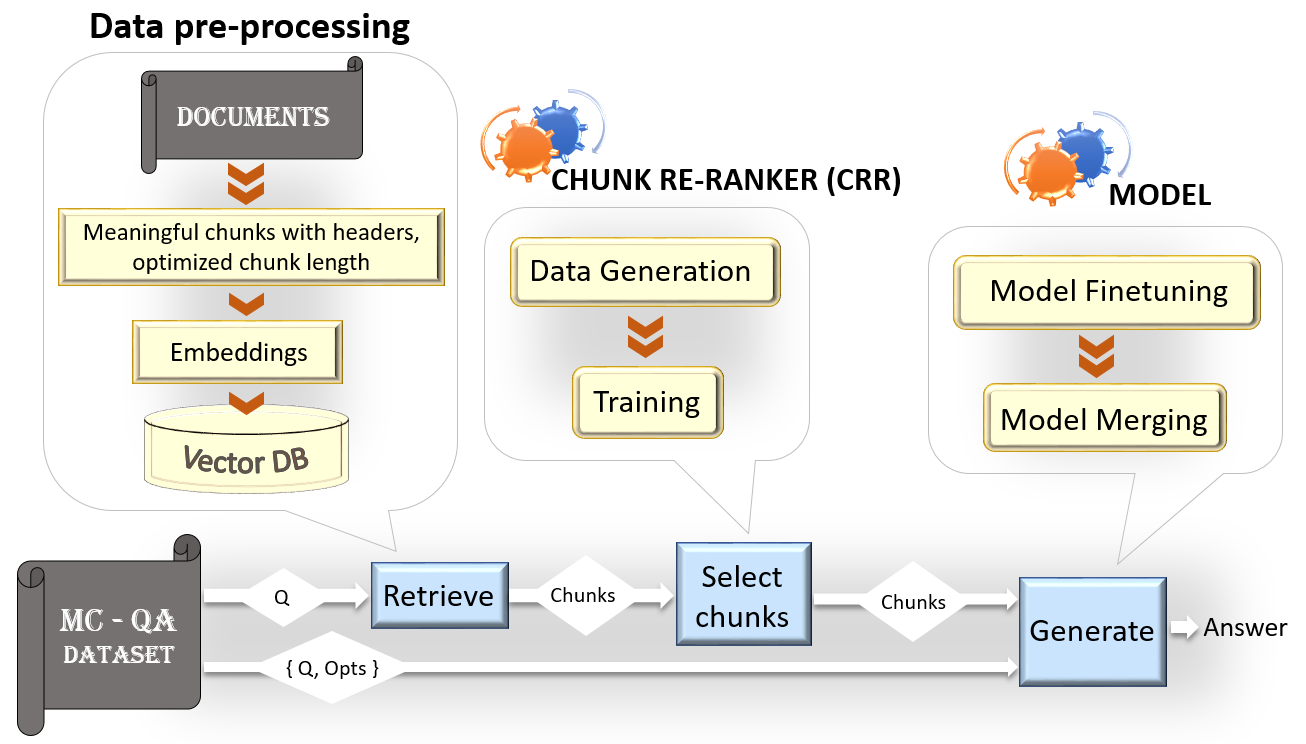}
    \caption{Our method workflow. The pipeline consists of the three main modules: 'Data pre-processing', 'CRR' and (SFT) 'Model'. In the 'Data pre-processing' module, document text is structured and cut into semantically meaningful chunks, followed by the vector DB creation. The 'CRR' module generates ground truth for consequitive training of the Chunk Re-Ranker Language Model. The rightmost 'Model' module performs pre-training and fine-tuning on the newly created data corpus and MC-QA set. During the inference stage (the bottom part of the scheme), the question is sent to the retriever to extract a set of chunks. Next, the CRR re-calculates the chunks relevance score with respect to the question. The most relevant chunks with the highest relevance score are then used as the context in the generation part of the workflow.}
    \label{fig:flow}
\end{figure}

Our approach comprises of several methodologies that contribute to the accuracy of the final system. These include: 1) LLM-friendly data representation with efficient text chunks creation; 2) Subsequent dedicated chunk re-ranker for context filtering; 3) Effective knowledge fusion of SFT models. We designed and implemented a corresponding pipeline that gradually improves the quality of MC-QA algorithm stages. We demonstrate that good results and generalization in MC-QA can be achieved even with limited number of parameters model fine-tuned on a small dataset. Fig.\ref{fig:flow} illustrates our entire data processing and training pipeline. It is designed as a combination of independent modules, allowing each module to be modified and verified separately. Below we provide a step-by-step description of our pipeline and detail the improvements related to each stage.

\subsection{Data Analysis and pre-processing}\label{Main_DataPrep}

\subsubsection{Dataset}
Our input data were derived from a subset of the recently published TeleQnA dataset \cite{b14} and contained:
\begin{itemize}
\item 550 doc files with various descriptions and technical specifications from the telecommunication field, including but not limited to 3GPP, IEEE, and ITU standards.
\item A set of MC-QA, having 1461 tuples: [question, answer options, answer, explanation, specification]. See Fig.\ref{fig:data_example} for data example.
\end{itemize}

In order to fully exploit the limited amount of the available data, and to be able to evaluate our results in a statistically meaningful manner, we used a cross-validation approach. In particular, the MC-QA dataset was divided into random splits, and training and evaluation were performed on each split (we provide more details in the following sections).

\subsubsection{Data pre-processing and structuring} 

\begin{figure}
    \centering
    \includegraphics[width=1\linewidth]{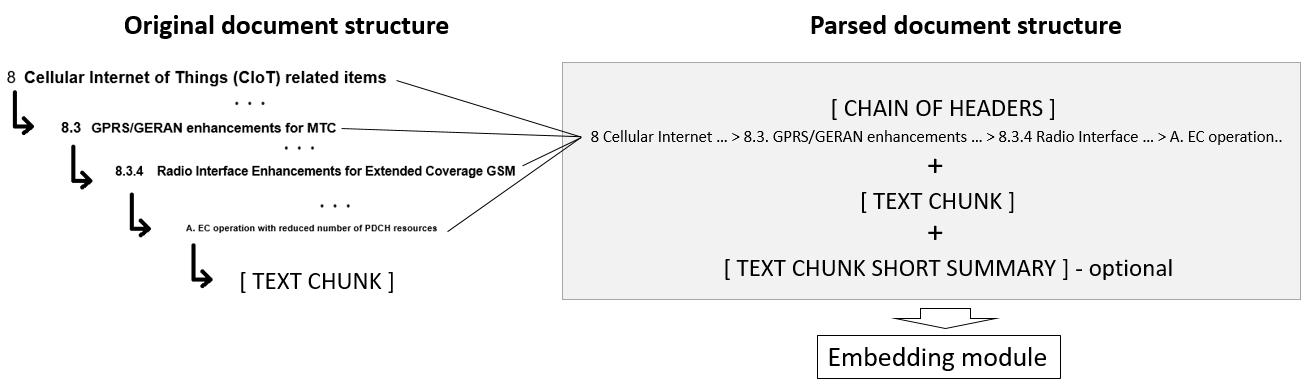}
    \caption{Data parsing example: left - the original unstructured text under the title of chapter, subtitles of a few subsequent chapters, followed by the original text chunk; right - a post-processed structure: each chunk is enriched with an added header comprising of document title, chapter title and subtitles.}
    \label{fig:data-parsing}
\end{figure}

It is well known that for training better models, the amount of training data is not as crucial as the data quality \cite{b4}, \cite{b5}, \cite{b16}, \cite{b17}.
High-quality context and MC-QA data ensures that the model develops a clear and focused understanding of the task. To ensure that LLM can effectively understand the text, the input is restructured into a standard form. All the tables and figures are extracted and stored separately. After we parse the input documents and convert them into the required format, each document is segmented into meaningful text chunks optimized both for training and retrieval purposes: each chunk corresponds to a chapter of the document, and the chunks are deduplicated. To provide additional context for each chunk, we maintain a chain of titles and sub-titles associated with each chunk (which is utilized both in the retrieval and in the training processes). The example of the original and of the parsed data chunks is shown in Fig.\ref{fig:data-parsing}. After the data corpus is divided into meaningful chunks with the chain of captions, we apply one of the recent embedding: \cite{b32} and \cite{b33} (referred further in the text as SFR-2 and MS embeddings), to convert each chunk into embedding vector, and build a vector data base. If the length of the chunk exceeds the pre-defined length, we split it into a few sub-chunks, keeping a chunk header for each sub-chunk. We also retain the relevant information regarding whether sub-chunks belong to the same larger chunk and their respective order.

\begin{figure}
    \centering
    \includegraphics[width=1\linewidth]{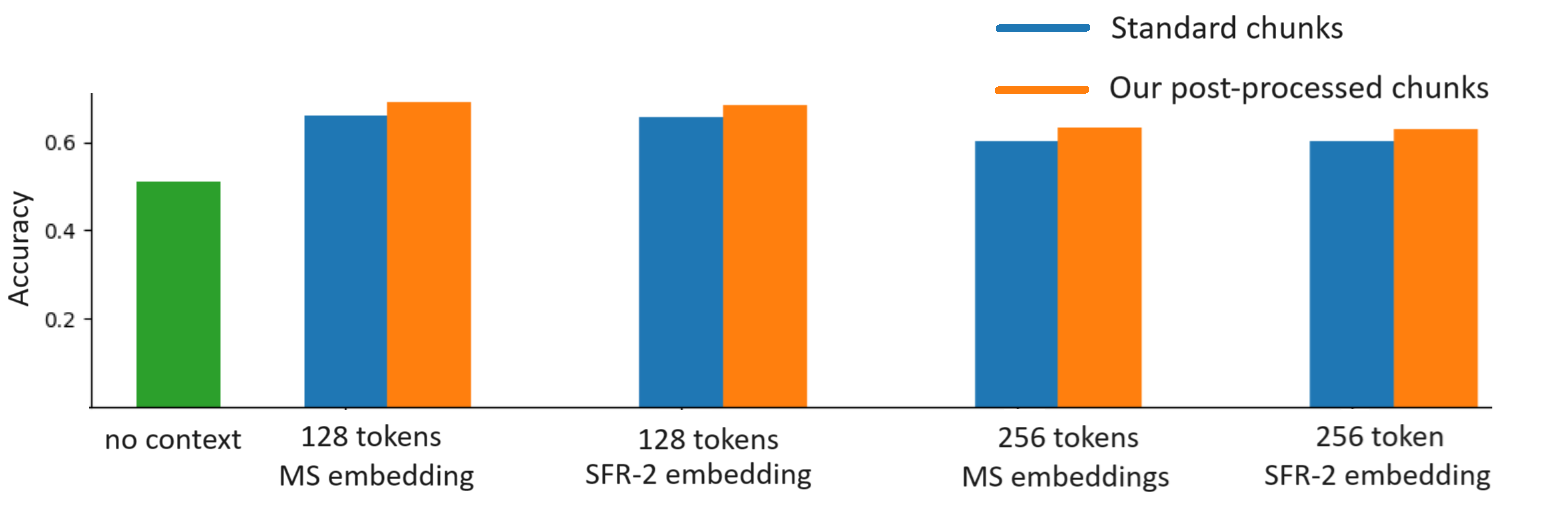}
     \caption{Influence of the context content on the accuracy of the model. The bar height encodes the model accuracy: no context (green), with 7 extracted standard chunks as the context (blue), and 7 extracted chunks using our text restructuring (orange). Prompting with the context based on our structured text chunks improves accuracy by up to 4\% (which constitutes a substantial improvement relative to the data pre-processing cost). We used two embedding types (MS and SFR-2); both yield similar accuracy.}
    \label{fig:old-new chunks}
\end{figure}

Fig.\ref{fig:old-new chunks} illustrates the effectiveness of this document organization method by comparing the MC-QA accuracy for the base Phi-2 model with the accuracies of the RAG-based approached using both standard chunks and our post-processed chunks. It is important to note, that Phi-2 model is highly sensitive to the input prompt, which may result in varying accuracy reports by different researchers. For the RAG-based approach we ensure that the same prompt was used across all types of input chunks.

This approach may increase the number of chunks, as not all sub-chapters reach the maximum allowable chunk length (128 and 192 tokens in our experiments). However, each chunk becomes more semantically rich and easier for the RAG-based model to process.

\subsubsection{Context extraction: data chunk size vs number of chunks}

Small language models, such as Phi-2, are limited in the maximum length of the context they can handle. This puts strong limitation on the number of retrieved documents and their length. We, therefore, need to find an optimal trade-off between the size of the retrieved chunks and the number of chunks. We can use the following two strategies:

\textit{Strategy 1}: We can increase the size of the retrieved text chunk, while decreasing the number of chunks; or 

\textit{Strategy 2}: To extract a larger number of smaller chunks. We provide a set of experiments to analyse how model accuracy depends on the number of chunks provided (for different chunk size), and the place of the 'golden' chunk. The chunk is labeled as the 'golden' for the given pair of question-answer, if two conditions hold:
\begin{itemize}
    \item the use of this only chunk as the context allows the model to answer the question correctly;
    \item the model is not able to answer correctly without this chunk.
\end{itemize}

\begin{figure}
    \centering
    \includegraphics[width=1\linewidth]{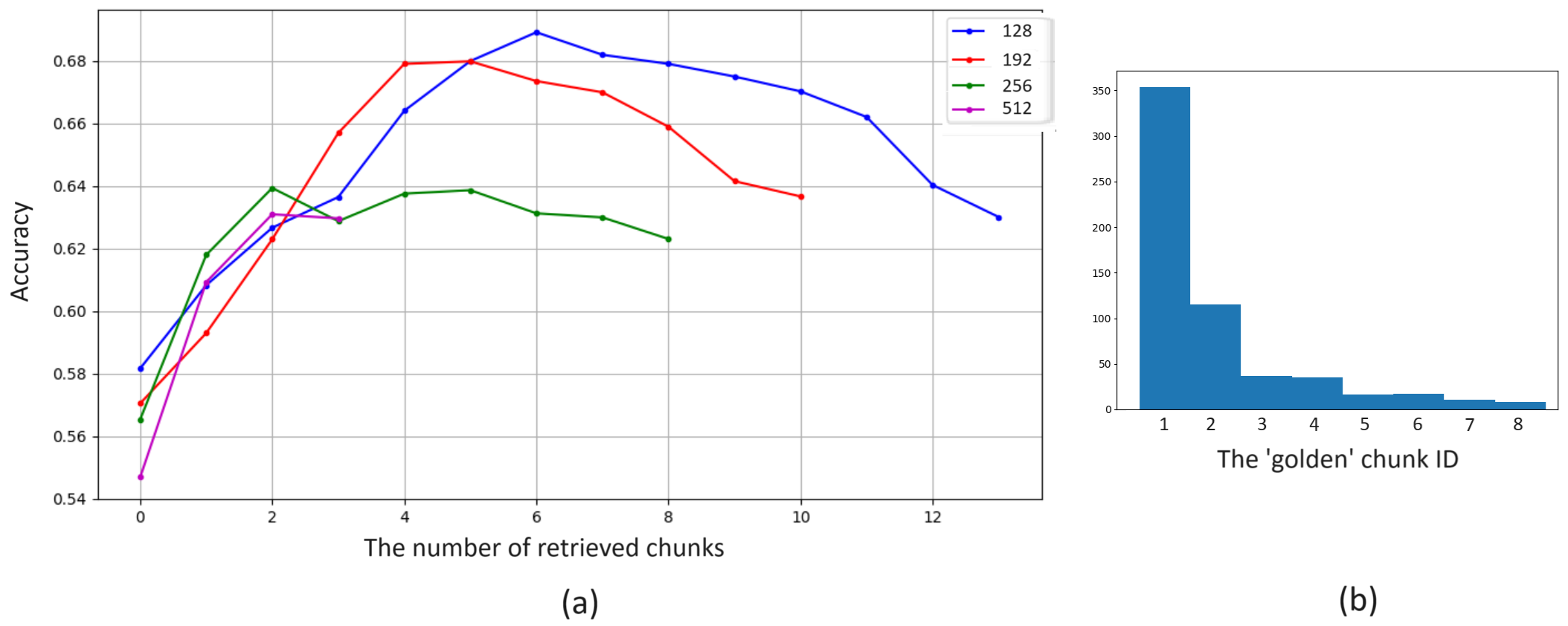}
    \caption{Chunk size vs. the number of chunks trade off, and  the golden chunk order: (a) Accuracy of the trained model as a function of the chunk size and the number of chunks used as the context prompt for the RAG (blue, red, green and magenta encode the chunk size of 128, 192, 256 and 512 tokens respectively); (b) The number of the 'golden' chunks vs their corresponding IDs: the 'golden' chunk is usually retrieved the first, but in many cases it appears in the 2-nd, 3-rd and even 7-th position out of total 7.}
    \label{fig: acc vs chunk size}
\end{figure}

Fig.\ref{fig: acc vs chunk size} (a) shows how the accuracy of the base model changes with respect to the number of chunks provided as context for RAG. We analyzed a few chunk sizes from 128 to 512. The choice of the chunk size is determined by the maximal context length of the model, which is 2048 tokens in our case. On the one hand, the model is very sensitive to the context length and to the position of the 'golden' chunk in the prompt (see our experiment described in  Section III-B), which suggests the use of the smallest number of chunks possible. On the other hand due to the imperfection of the retrieval and embedding models, that don't take into account the nuances of domain specific terminology, the 'golden' chunk is not always the first chunk to be selected, see Fig.\ref{fig: acc vs chunk size} (b). In some cases the place of the 'golden' chunk in the retrieved sequence is the last one (7 out of 7). 
The experiments show the \textit{Strategy 2} is preferable in our case.

\subsection{Chunks Re-Ranking and filtering}

In our approach, we utilize in-context LLM predictions. Specifically, we first identify several relevant text chunks from the data corpus, in response to a given question.  The LLM performance is affected by the total number of retrieved chunks, their particular order, and the number of irrelevant retrieved chunks. There some recent approaches deploying post-processing steps to improve the quality of the RAG-based techniques, such as removing unnecessary chunks, re-ranking them, adding chunk summarization and others. Although, approaches that are effective for large models, may not be applicable for small ones (see e.g. \cite{b28}). Also we were forbidden to use other LLMs as a part of a pipeline. We performed a small study of how sensitive Phi-2 is to these parameters. We selected 1200 meaningful text chunks from provided data corpus and run Mixtral-8x7B \cite{b34} model to generate 1200 MC-QA from selected chunks. Although generated questions' quality is inferior to the quality of human generated questions (we didn't use these generated questions to augment MC-QA training data), this controlled experiment provided us with a set of 'golden' chunks and allowed to analyze the impact of the position of the 'golden' chunk in the context. The observations are:
\begin{itemize}
    \item The inclusion of a single random chunk to the 'golden' chunk decreases accuracy by up to 2\%. 
    \item The inclusion of 6 random chunks decreases accuracy up to 4-5\%
\end{itemize}

 The strategy we have chosen according to this toy experiment, was to retrieve the reasonable number of chunks and then to try to filter them. Finally we found out that for Phi-2 use of a domain-specific Chunk Re-Ranker (CRR) and filtering significantly improves the final accuracy. In particular, we train the CRR (based on Phi-2 model) to assign a score to each retrieved chunk and use this score instead of the score provided by the Retriever to select the context chunks. The score indicates how well a specific chunk contributes to the correct answering to a given question. 

In order to create the dataset for CRR training, we used Mixtral-8x7B - a more powerful off-the-shelf LLM with strong reasoning capabilities. Given a set of retrieved chunks, our CRR assigns a discrete score from '1' to '5' to each chunk, with a '1' indicating the chunk does not contribute to answering the question, and a '5' indicating the chunk contains all the necessary information to fully address the question. Only chunks with the scores of '4' and '5' are retained for the context, while all others are discarded. In the case wherein no chunks got a score of '4' or '5', the context is created from the three original chunks ordered as they came from the retrieval.

\subsection{Model fine-tuning and merging}
We performed an initial model pre-training on the data corpus parsed with respect to the meaningful chunks. Then a pre-trained model was fine-tuned on MC-QA data with context obtained from the retrieval, followed by the reranking procedure.

\textit{Cross-validation.} In line with the principles of cross-validation and given the limited training data, we performed model fine-tuning on each random split. The size of the validation set for each split was adjusted to the size of the private test set (336 questions). Thanks to the small model size (2.5B) the SFT process was completed in just a few minutes (about 15 minutes, 3 epochs, 2 GPUs) making a cross-validation procedure - where the model is retrained and evaluated multiple times - feasible. By evaluating the model on unseen portions of the data during each iteration, we gained valuable insights into the data distribution and the model's generalization capabilities.

\textit{Prompting strategy and augmentations.} Our choice was to deploy the multiple choice prompting (MCP) strategy. For the MC-QA approach to be effective, the model must be able to associate answer options with the corresponding symbols representing them. Furthermore, the model needs to have the multiple choice symbol binding (MCSB) ability. In particular, to ensure this ability, we train the model both with the default symbols '1/2/3/4', and with the default symbols replaced by 'A/B/C/D'. Any other additional replacements with reasonable alternatives, such as 'a/b/c/d' brought no significant improvement. Furthermore, removing option ID's and requesting the model to select option contents directly have not brought any improvements, probably due to the model's limited size.

\textit{Data balancing.} To mitigate selection bias, we conducted a permutation of the option contents. We shuffled the answer options such that the distribution of the correct answer is uniform across 5 options.

\textit{Multiple models fusion.}
To increase model generalization ability, and to improve its robustness, we tried different model merging methods. First, we merged the weights of the models fine-tuned separately on numbers and on capital letters ground truth, associated with the answer options. Such fusion improved the performance of our final model, as compared to both separately trained models (see Fig.\ref{fig:7 splits accuracy})
Next, in order to be able to extract the knowledge from the whole MC-QA dataset, and, at the same time, to be able to test the models on validation sets, we combined SFT models obtained from each cross-validation split. In particular, we adopted similar approach of merging model weights from models with identical structures, but obtained through different strategies or configurations - having the same numbers of parameters in each layer, the merge models merges the two models by linear averaging.  This further improved overall performance on the limited dataset by 2\%. The weights of all the SFT models from all the splits were merged into a single new model. The use of the weighted average to let each model to contribute non equally, for example according to initial accuracy didn't show an improvement. Our model merging strategy showed to be superior to the traditional technique of model ensemble (combining the outputs of multiple models to enhance overall system performance).

\section{Experiments and Implementation details}

\textit{Data.}  The documents were divided into chunks as explained in Section Dataset (about 127,000 chunks). A typical sequence window for an embedding model is 512 tokens, which also makes a practical target for chunk size, but we made them smaller (128-192) to balance between the number of chunks and the maximal context window of Phi-2 (2048 tokens). 

We split the original MS Word documents according to the internal document structure where any titles with a "headings" font serve as a text separator. If any of the resulting chunks exceeds 128 (192) characters, we further split it into smaller pieces using a sliding window of size 128 (192)  and stride 64. We add a small header to each chunk which contains the path to the text within the document Table of Contents tree. 

Although all the documents were used for the initial pre-training, the MC-QA dataset was divided into 7 random splits: for each split we left 336 questions for evaluation (the size of the unpublished public test set), and the rest were used for the SFT.

\textit{CRR. } We fine-tune a Phi-2 model on the generated rankings dataset. The dataset contains 1095 triplet samples: {chunk, question, rank}. The Re-Ranker was optimized using the standard next-token prediction objective, with the target text being a single digit representing the correct rank. For the training, we use batch size of 64, Adam optimizer with learning rate of 5e-6 and a constant scheduler with 70 warm-up steps. The ranker was trained for total of 100 epochs. After CRR training was completed, the final model was fine-tuned on the seven MC-QA splits of the training data with the CRR chunk filtering. The CRR improved the final accuracy by about 3\% on average.

\begin{figure}
    \centering
    \includegraphics[width=1\linewidth]{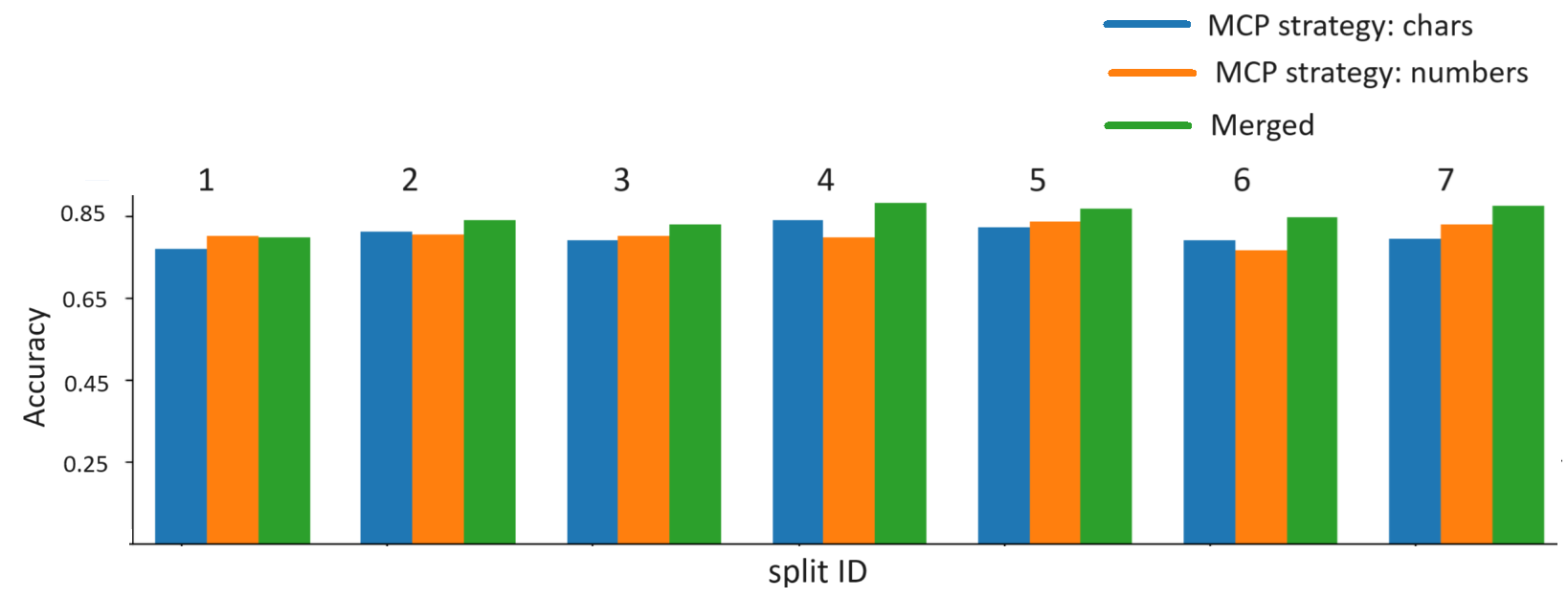}
    \caption{The accuracy of the SFT model on seven random data splits. For each split we fine-tuned and evaluated the model using one of the MCP strategies: with chars (A/B etc shown in blue) and numbers (1/2 etc shown in orange) encoding the answer options. For each split two fine-tuned models were merged to get a new single model (green). As can be seen, merged model outperforms both MCP models on almost all the splits.}
    \label{fig:7 splits accuracy}
\end{figure}

\textit{SFT model. } We estimate SFT model on each of 7 splits: chars, numbers and merged model. Then we merged all the 7 models into a single one. Fig.\ref{fig:7 splits accuracy} shows 7 bar triplets corresponding to the chars, numbers and merged model (colors). Merged weights were tested on char-based options. Merging improves char/number models by 2.5-8\%
The resulting merged model was evaluated on public (336 samples) and private (2,000 samples) test sets and achieved accuracy of  77\% and 79.7\% respectively. Some of the models that wasn't chosen for the final submission exceeded 81\% on the private dataset. 

 We used batch size of 16, Adam optimizer with learning rate 1e-5 and a polynomial scheduler with 10 warm-up steps. The model was trained for total of 3 epochs.

\section{Conclusion}
In this paper we demonstrated that, by improving data quality at each stage of the LM training pipeline, it is possible to achieve high downstream task accuracy, even with small LM, such as Phi-2. We proposed and detailed several such techniques, that are tailored specifically for small LM training. These techniques include: targeted data pre-processing for more efficient knowledge digestion by the LM; text chunk re-ranking for more informative context extraction; multiple data subsets model merging for better generalization. Corresponding experiments confirm the efficacy of the proposed techniques.

\bibliographystyle{unsrt}  
\bibliography{main} 

\end{document}